\documentclass[letterpaper]{article} 
\usepackage{aaai2026}  
\usepackage{times}  
\usepackage{helvet}  
\usepackage{courier}  
\usepackage[hyphens]{url}  
\usepackage{graphicx} 
\usepackage{natbib}  
\usepackage{caption} 
\usepackage{algorithm}
\usepackage{algorithmic}
\usepackage{amsmath}
\usepackage{amssymb}
\usepackage{booktabs}
\usepackage{multirow}
\usepackage{multicol}
\usepackage{array}
\usepackage{makecell}
\usepackage{xcolor}
\usepackage[table]{xcolor}
\usepackage{pdfpages}
\usepackage{newfloat}
\usepackage{listings}
\DeclareCaptionStyle{ruled}{labelfont=normalfont,labelsep=colon,strut=off} 
\floatstyle{ruled}
\newfloat{listing}{tb}{lst}{}
\floatname{listing}{Listing}
\title{SOMA: Feature Gradient Enhanced Affine-Flow Matching \\for SAR-Optical Registration}
\author{
    Haodong Wang\textsuperscript{\rm 1,2,3}\equalcontrib,
    Tao Zhuo\textsuperscript{\rm 4}\equalcontrib\\
    Xiuwei Zhang\textsuperscript{\rm 1,2,3}\thanks{Corresponding authors.},
    Hanlin Yin\textsuperscript{\rm 1,2,3}\footnotemark[2],
    Wencong Wu\textsuperscript{\rm 1,2,3},
    Yanning Zhang\textsuperscript{\rm 1,2,3}
}
\affiliations{
    \textsuperscript{\rm 1}School of Computer Science, Northwestern Polytechnical University\\
    \textsuperscript{\rm 2}Shaanxi Provincial Key Lab. of Speech and Image Information Processing\\
    \textsuperscript{\rm 3}National Engineering Laboratory for Integrated Aero-Space-Ground-Ocean Big Data Application Technology\\
    \textsuperscript{\rm 4}College of Information Engineering, Northwest A\&F University\\

    traslauc@mail.nwpu.edu.cn, xwzhang@nwpu.edu.cn, iverlon1987@nwpu.edu.cn

}

\begin{document}

\maketitle

\begin{abstract}
Achieving pixel-level registration between SAR and optical images remains a challenging task due to their fundamentally different imaging mechanisms and visual characteristics. Although deep learning has achieved great success in many cross-modal tasks, its performance on SAR-Optical registration tasks is still unsatisfactory. Gradient-based information has traditionally played a crucial role in handcrafted descriptors by highlighting structural differences. However, such gradient cues have not been effectively leveraged in deep learning frameworks for SAR-Optical image matching. To address this gap, we propose SOMA, a dense registration framework that integrates structural gradient priors into deep features and refines alignment through a hybrid matching strategy. Specifically, we introduce the Feature Gradient Enhancer (FGE), which embeds multi-scale, multi-directional gradient filters into the feature space using attention and reconstruction mechanisms to boost feature distinctiveness. Furthermore, we propose the Global-Local Affine-Flow Matcher (GLAM), which combines affine transformation and flow-based refinement within a coarse-to-fine architecture to ensure both structural consistency and local accuracy. Experimental results demonstrate that SOMA significantly improves registration precision, increasing the CMR@1px by 12.29\% on the SEN1-2 dataset and 18.50\% on the GFGE\_SO dataset. In addition, SOMA exhibits strong robustness and generalizes well across diverse scenes and resolutions.

\end{abstract}

\begin{links}
    \link{Code}{https://github.com/traslauc/SOMA}
\end{links}

\section{Introduction}

\begin{figure}[t]
    \centering
        \includegraphics[width=0.46\textwidth]{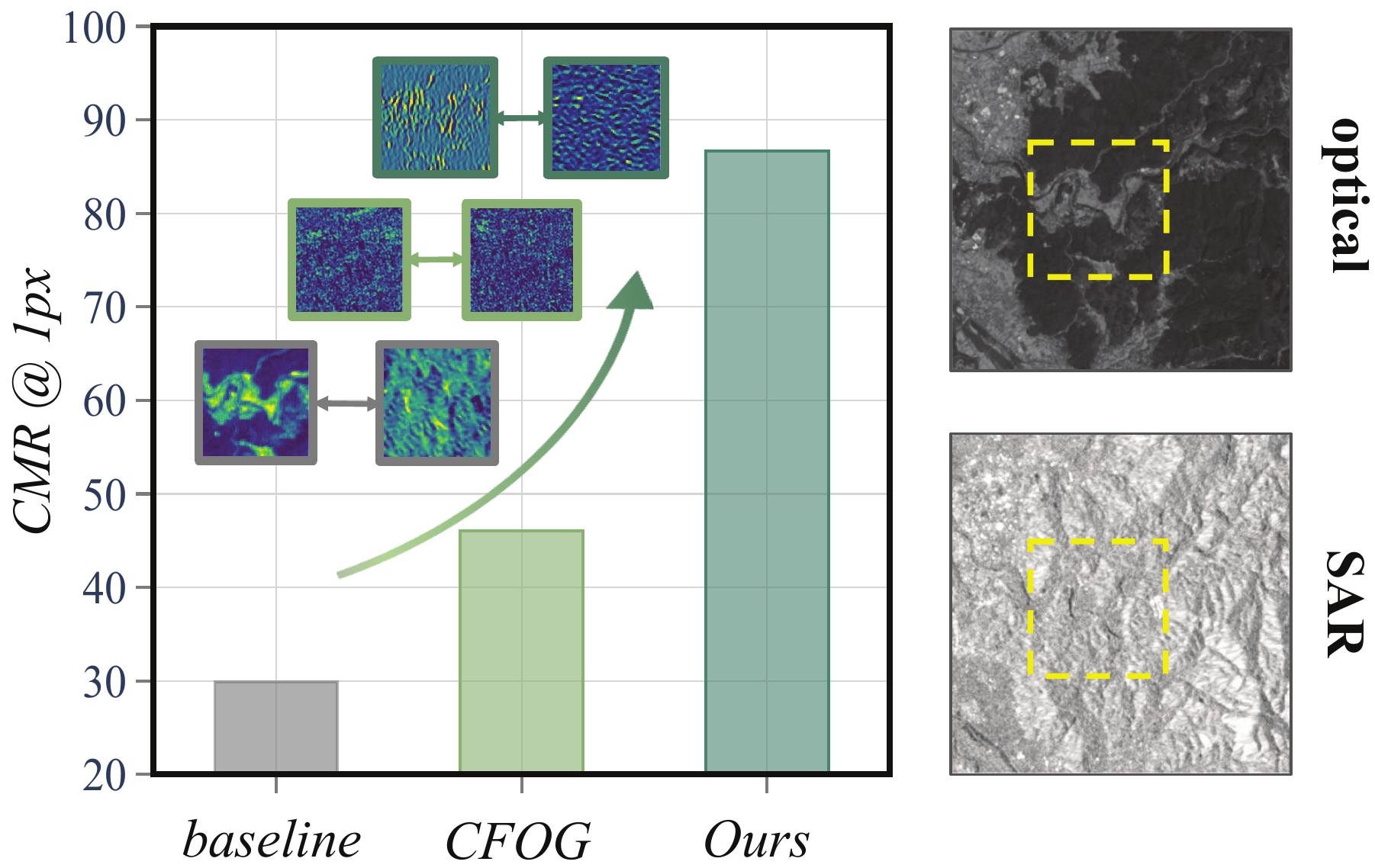}
	\caption{\textbf{Comparison of CMR@1px} among methods leveraging different types of feature representations. The proposed SOMA, which incorporates feature gradient enhancement, significantly outperforms both the conventional CNN baseline and the CFOG that leverages image gradients.
}
	\label{fig:intro}
\end{figure}

Multi-modal remote sensing image registration is a fundamental step in multi-source data fusion and analysis, supporting tasks such as object detection, change detection, 3D reconstruction, and joint classification \cite{jiang2021review}. Among various modalities, Synthetic Aperture Radar (SAR) and optical imagery have received particular attention due to their complementary sensing characteristics. SAR provides all-weather, all-day imaging with strong sensitivity to surface structure and roughness, even through cloud cover. Optical imagery, in contrast, offers high-resolution visual details and semantically rich information. Their combination is especially valuable in emergency response and terrain interpretation under adverse weather. Accurate alignment between SAR and optical images is thus essential for effective joint analysis.

Recent years, learning-based SAR-Optical image registration methods have demonstrated remarkable improvement in robustness~\cite{deepdes,deepfea,li2023multimodal}. However, in practical applications, the substantial disparity in imaging mechanisms and image characteristics between SAR and optical images presents a significant challenge for accurate registration. In particular, SAR images often contain regions severely affected by nonlinear radiometric distortions and speckle noise, where the pixel-level features lack distinctiveness from their surroundings. This makes it difficult for existing methods to sufficiently extract reliable dense features, ultimately limiting the following matching precision.

In fact, it is observed that gradient information plays a critical role in the design of handcrafted features. Handcrafted features \cite{CoFSM,zhang2024robust} utilize multi-directional and multi-scale gradient extraction strategies to effectively capture structural information, allowing local differences to be distinguished from their surroundings. However, when gradient extraction is directly applied to raw images, these features are easily overwhelmed by noise, leading to unstable responses and disturbed semantic information, as shown in Figure~\ref{fig:intro}.

\begin{figure}[t]
    \centering
        \includegraphics[width=0.47\textwidth]{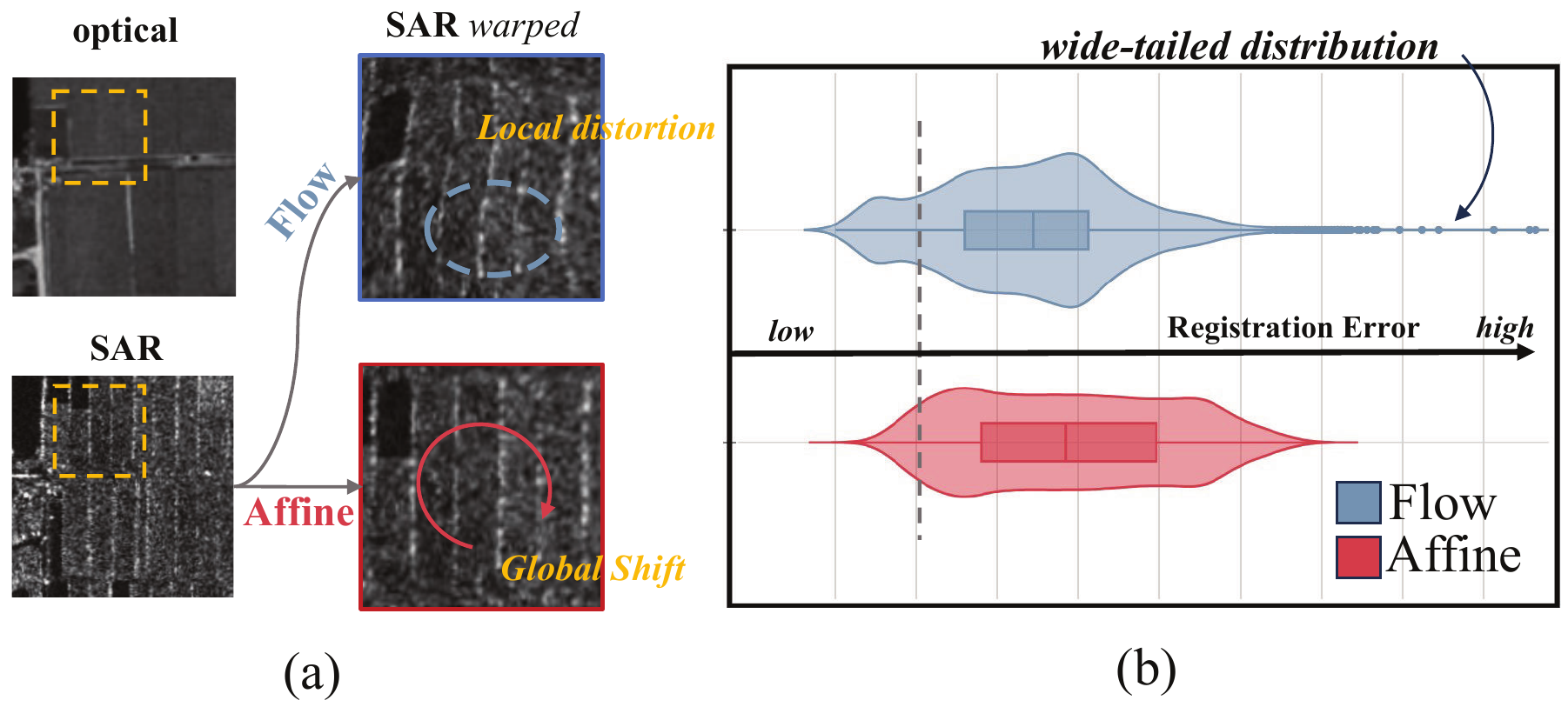}
	\caption{\textbf{(a)} Flow fields prioritize local warp, but leave distortion. Affine fields maintain structural consistency via global transforms but lost local precision. \textbf{(b)} Violin plots further reveal distinct error profiles.
    } 
    \label{fig:diff}
\end{figure}

Building on this analysis, we explore the explicit integration of multi-directional and multi-scale gradient features into learning-based feature extraction process. Unlike traditional methods that compute gradients directly from images, we apply directional gradient filters to the feature pyramid to better capture structural cues. However, high-level abstract features often contain redundant and entangled information across spatial and channel dimensions. Direct gradient filtering in such feature spaces can severely degrade feature integrity.
To address this, we adopt a feature reconstruction mechanism~\cite{SCC} that reduces redundancy in spatial and channel dimensions, retaining features crucial for effective gradient filtering and accurate correspondence estimation.
We further incorporate the attention mechanism and dilated convolutions to effectively integrate and reinforce the extracted gradient cues. 
Additionally, a final Gaussian smoothing step is applied to reduce early residual noise. We encapsulate this gradient enhancement into a dedicated module named Feature Gradient Enhancer (FGE).

Although FGE enhances the deep features, achieving accurate dense registration still requires a robust matcher. Most existing multimodal remote sensing image matching approaches focus on learning a single geometric transformation. However, global transformation lacks local precision, while dense warp tends to compromise structural consistency. As a result, these methods may suffer from local distortion under challenging conditions, as shown in Figure~\ref{fig:diff}.

We propose a Global-Local Affine-Flow Matcher (GLAM) to fix this issue, which introduces affine regression in a coarse-to-fine manner. The affine regression captures large-scale structural alignment and provides consistent deformation priors, while the flow regression refines local details at progressively finer resolutions. By coupling these two transformations, our GLAM mitigates local misalignment and enhances global stability, leading to more precise and robust cross-modal matching.

Finally, to give the entire pipeline a stable coarse warp, we adopt a frozen DINOv2~\cite{dinov2,dinov2reg} vision backbone as the coarse-level feature encoder, inspired by RoMa~\cite{roma}. 
This global perceptual context from DINOv2 complements the FGE, providing reliable input to GLAM.

To this end, we propose a dense matching framework for SAR and optical image registration, termed SOMA (\textbf{S}AR-\textbf{O}ptical \textbf{MA}tching), which integrates the proposed FGE and GLAM modules within an end-to-end architecture to achieve robust and accurate SAR-Optical image registration.

Our main contributions are summarized as follows:

\begin{itemize}
    \item We propose FGE, a feature gradient enhancer that leverages multi-scale, multi-directional gradients to model distinctive representations, improving matching precision and robustness between SAR and Optical images.
    \item We propose GLAM, a global-local affine-flow matcher that forms a bidirectional coupling between local flow and global affine estimation, improving local accuracy while avoiding distortion.
    \item We introduce a frozen DINOv2 to stable init alignment and improve robustness in registration.
    \item We benchmark SOMA on diverse public datasets and it consistently surpasses state-of-the-art baselines, boosting CMR@1px by \textbf{12.29\%} on SEN1-2 and \textbf{18.50\%} on GFGE\_SO dataset, while retaining strong robustness and generalization in challenging scenes.

\end{itemize}

\begin{figure*}[t]   
	\centering
	\includegraphics[width=\linewidth,scale=1.00]{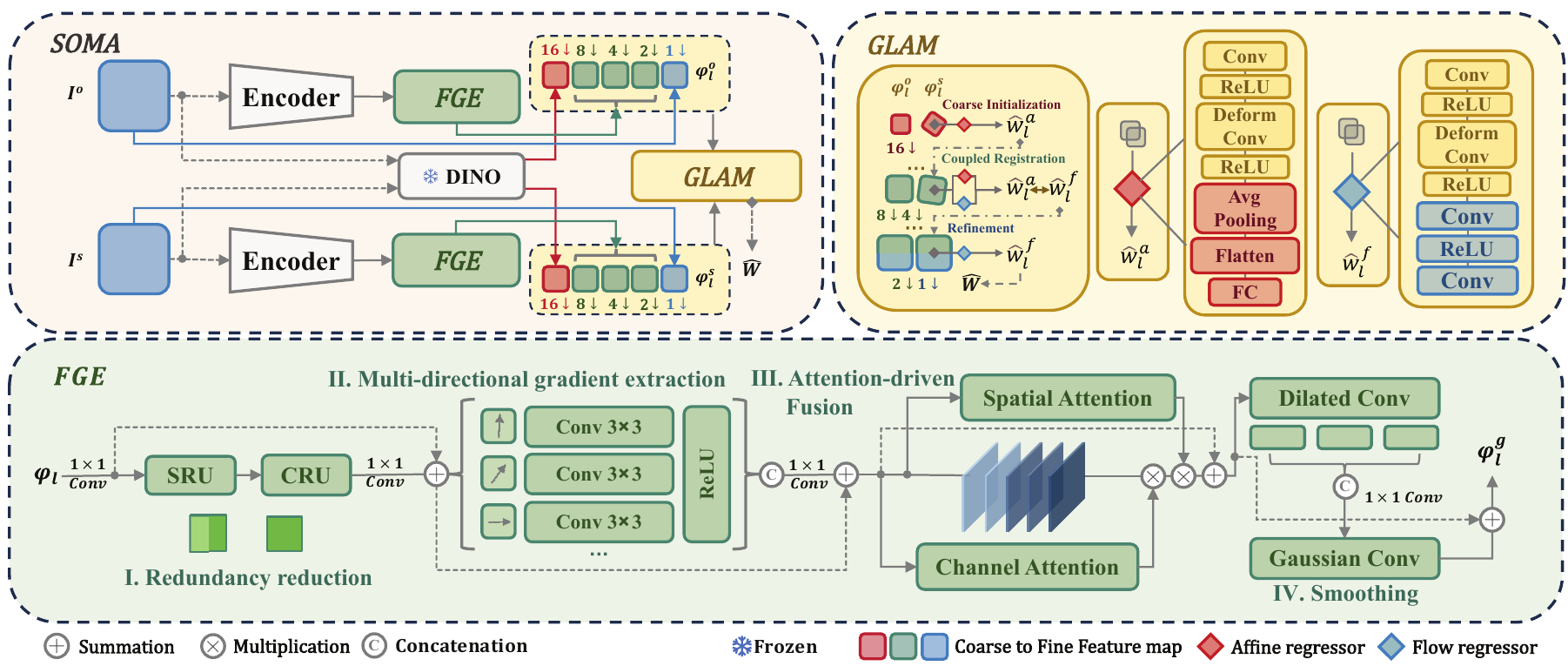}
	\caption{\textbf{SOMA framework with FGE and GLAM.}
                Given a SAR-Optical image pair, SOMA extracts multi-scale features by two separated ResNet50 encoders and a frozen DINOv2 branch, then enhances representations with Feature Gradient Enhancer (FGE), and performs hierarchical alignment using Global-Local Affine-Flow Matcher (GLAM). All convolution layers use a kernel size of $3\times3$ by default, unless stated otherwise.}
	\label{fig:framework}
\end{figure*}

\section{Related Work}

\subsection{SAR-Optical Registration Methods}

SAR-Optical image registration approaches can be broadly categorized into handcrafted methods and deep learning-based models. Handcrafted methods typically employ statistical similarity measures~\cite{hoss,mlss} or gradient-based descriptors such as CFOG and SAR-SIFT~\cite{sarsift,xiang2018sift,CoFSM,zhang2024robust} to extract structural cues. While these methods are effective in capturing geometry, they often suffer from limited robustness when confronted with large modality gaps in texture and radiometry. In contrast, recent learning-based methods leverage Siamese networks~\cite{liu2023optical}, attention mechanisms~\cite{zhang2021explore,deepfea}, or unsupervised learning strategies~\cite{munet} to extract modality-invariant features, improving generalization across diverse scenarios. 
However, such learning-based methods may still underperform in terms of fine-grained matching accuracy.
This limitation has also been noted in prior works~\cite{zhang2020ddfn}, suggesting the importance of highly distinctive feature representations for accurate pixel-level alignment.

\subsection{Pretrained Vision Models}

Robust cross-modal matching hinges on strong, transferable feature representations. Large self-supervised vision models have made this feasible: models such as DINOv2~\cite{dinov2}, CLIP~\cite{clip}, and iBOT~\cite{zhou2021ibot}—trained with contrastive or reconstruction objectives on massive image data—show remarkable generalization~\cite{dinoisbest}. Recent work across a wide range of vision tasks further confirms that keeping these models frozen boosts cross-domain performance: RoMa~\cite{roma} and OmniGlue~\cite{omniglue} exploit frozen DINOv2 features to stabilize semantic correspondences, while ChangeCLIP~\cite{dong2024changeclip} achieves comparable gains with CLIP embeddings. Nevertheless, effectively utilizing their representations for robust initialization in SAR-Optical matching remains an open issue. Motivated by this, we adopt a frozen DINOv2 backbone as an encoder component in our registration framework, providing a reliable foundation for subsequent coarse-to-fine alignment.

\subsection{Sparse and Dense, Local and Global Deformation Estimation}

Within multi-modal image matching, traditional keypoint-based methods are effective but often struggle in textureless or repetitive regions due to sparse and unstable correspondences.
Dense matching approaches, which estimate pixel-wise deformations, offer higher precision but lack global constraints, making them prone to structural inconsistencies~\cite{melekhov2019dgc,glunet}.
Affine transformation~\cite{affine} methods~\cite{murf,Xu_2022_CVPR} provide efficient global alignment but are often limited by parameter coupling and sensitivity to local variations.
To overcome these limitations, our GLAM establishes a cooperative optimization between global affine estimation and local flow refinement, achieving both structural consistency and fine-grained adaptability across diverse scenarios. 

\section{Method}
\subsection{Framework Overview}

As illustrated in Figure~\ref{fig:framework}, SOMA predicts a dense deformation field \(\hat{W} \in \mathbb{R}^{H \times W \times 2}\) that warps an optical image \(I^{\mathrm{o}} \in \mathbb{R}^{H \times W}\) to align with a SAR image \(I^{\mathrm{s}} \in \mathbb{R}^{H \times W}\). 

Dual encoders respectively extract five-level feature pyramids \(\varphi^{\mathrm{o}}_l, \varphi^{\mathrm{s}}_l \in \mathbb{R}^{C \times H_l \times W_l}\), where \(l \in \{1, 2, 4, 8, 16\}\) denotes the downsampling factor with respect to the original resolution. At the coarsest scale (\(l=16\)), a frozen DINOv2 is incorporated for robust features. At intermediate scales, we apply the Feature Gradient Enhancer (FGE) to explicitly inject multi-directional gradient into the feature space. FGE first reduces spatial and channel redundancy, then applies fixed directional filters followed by attention-based fusion and Gaussian smoothing, yielding enhanced features with improved structural distinctiveness.

These features are fed into the Global-Local Affine-Flow Matcher (GLAM), which estimates a hierarchy of deformation fields \(\hat{W}_l\). In a coarse-to-fine manner, an affine regressor predicts global transformation parameters \(\theta_l \in \mathbb{R}^{2 \times 3}\), which are then converted into a full-resolution displacement field \(\hat{W}_l^a \in \mathbb{R}^{H_l \times W_l \times 2}\), while a flow regressor estimates residual flow \(\hat{W}_l^f \in \mathbb{R}^{H_l \times W_l \times 2}\), refining the warped result from the previous level. In addition, a pixel-wise certainty map \(\hat{p} \in \mathbb{R}^{H \times W}\) is predicted at the finest level to weight supervision during training and is omitted during inference.

\subsection{Feature Extraction}

We extract multi-scale features at four levels \(l\!\in\!\{1,2,4,8\}\) using two modality-specific ResNet50 backbones trained from scratch. However, deep CNN features are sensitive to modality imbalance: for instance, SAR images may exhibit severe sudden intensity
changes in specular or noisy regions, leading to unstable responses~\cite{psic}.

To stabilize coarse alignment, we incorporate a frozen DINOv2 at \(l=16\). Pretrained via self-distillation, DINOv2 produces contrast-aware embeddings that remain perceptually meaningful across modalities~\cite{dinoisbest}. These features serve as a robust structural anchor at the coarsest level, helping the hierarchical matcher establish reliable correspondences before refinement at higher resolutions.

\subsection{Feature Gradient Enhancer (FGE)}

Accurate fine-scale registration requires reliable feature description ability with high distinctiveness across SAR and optical modalities. Although gradient cues are naturally suited for capturing such representation, directly applying fixed gradient filters to high-dimensional CNN feature maps often amplifies aliased noise~\cite{aliased}. Moreover, how to effectively integrate gradient information into deep neural networks remains an open issue. To address this, we propose the Feature Gradient Enhancer (FGE), which first reduces feature redundancy, and then explicitly integrates multi-directional gradient cues using attention and dilated convolutions, resulting in robust and distinctive feature representations. The details of FGE’s four main components are elaborated as follows.

\paragraph{Redundancy reduction.}  
FGE begins by reconstructing the input feature map \(\varphi_l\). We adapt two lightweight modules from SCConv~\cite{SCC}: a spatial reconstruction unit \(\mathrm{SRU}(\cdot)\) and a channel reconstruction unit \(\mathrm{CRU}(\cdot)\), suppressing noisy and aliased responses.
The reconstructed feature $F_{\mathrm{recon}}$ serves as a foundation that facilitates effective gradient extraction:
\begin{equation}
F_{\mathrm{recon}} = \mathrm{CRU} \bigl( \mathrm{SRU}(\varphi_l) \bigr)+\varphi_l .
\label{eq:feature_recon}
\end{equation}
Residual connection is employed to maintain stable information throughout the reduction process and subsequent enhancement stages.

\paragraph{Multi-directional gradient extraction.}  
We apply a bank of rotated Sobel-like kernels $\{K_{\theta_i}\}$ on $F_{\mathrm{recon}}$, each capturing a directional gradient: 
\begin{equation}
G_i = \mathrm{ReLU}\bigl(K_{\theta_i} * F_{\mathrm{recon}}\bigr),\quad i=1,\dots,N.
\end{equation}
In practice, we adopt $N=8$ different directions with $\theta_i \in \{0^\circ, 22.5^\circ, 45^\circ, 67.5^\circ, 90^\circ, 112.5^\circ, 135^\circ, 157.5^\circ\}$. The outputs $G_i$ of these \(N\) branches are concatenated and fused into $F_{\mathrm{grad}}$ via \(1\times1\) convolution,
\begin{equation}
F_{\mathrm{grad}} = \mathrm{Conv}_{1\times1}\bigl([G_1,\dots,G_N]\bigr)+F_{\mathrm{recon}}.
\end{equation}

\paragraph{Attention-driven fusion.}  
To adaptively enhance the most informative gradient cues under varying scene conditions, we introduce simple attentions \cite{cbam} following the directional filtering. Specifically, we apply a lightweight channel-attention block \(\mathrm{CA}(\cdot)\) and a spatial-attention block \(\mathrm{SA}(\cdot)\) to selectively emphasize salient gradient responses. This mechanism compensates for the fixed nature of gradient kernels and enhances the discriminability of structural features crucial for matching.
Formally, the attention-enhanced feature $F_{\mathrm{att}}$ is computed as:
\begin{equation}
F_{\mathrm{att}} = F_{\mathrm{grad}} \otimes \mathrm{CA}(F_{\mathrm{grad}}) \otimes \mathrm{SA}(F_{\mathrm{grad}}) + F_{\mathrm{grad}},
\end{equation}
where \(\otimes\) denotes element-wise multiplication.

To enrich structural representation with multi-scale context, we apply dilated convolutions with dilation rates \(\{1, 2, 3\}\) to \(F_{\mathrm{att}}\). The resulting feature maps $F_{\mathrm{ms}}$ are fused via \(1\times1\) convolution as well:
\begin{equation}
F_{\mathrm{ms}} = \mathrm{Conv}_{1\times1}\bigl([\mathrm{Dilate}_d(F_{\mathrm{att}})]_{d=1}^3\bigr).
\end{equation}

\paragraph{Smoothing.}
To stabilize the early stages of training, we apply a depthwise convolution initialized with a Gaussian low-pass kernel $\mathrm{Gauss(\cdot)}$. The final enhanced feature map $\varphi_{l}^{g}$ is obtained by adding the attention-enhanced feature:
\begin{equation}
\varphi_{l}^{g} = F_{\mathrm{att}} + \mathrm{Gauss}\bigl(F_{\mathrm{ms}}\bigr).
\end{equation}

FGE is applied to the levels \(l\in\{2,4,8\}\). Its output $\varphi_{l}^{g}$ retains the same shape as \(\varphi_l\) and is directly consumed by the hierarchical matcher. For consistency of notation, we hereafter still denote \(\varphi^g_l\) as \(\varphi_l\).

In addition, how the proposed FGE affects cross-modal feature matchability is discussed in the Supplement. Interestingly, we observe that the feature gradient enhancement leads to a notable decrease in cosine similarity between corresponding SAR and optical features, suggesting that high similarity is not a necessary condition for effective cross-modal matching.

\subsection{Global-Local Affine-Flow Matcher (GLAM)}
\label{sec:glam}
Most existing multimodal remote sensing image matching approaches rely on regressing a single geometric transformation to align cross-modal inputs. However, due to the inherent modality gap between SAR and optical images, relying solely on either global or local transformations often proves insufficient. 

GLAM estimates two complementary deformation fields, an affine field $\hat{{W}}^{\mathrm{a}}_l$ and a flow field $\hat{{W}}^{\mathrm{f}}_l$, which further correct local misalignments remaining from the previous level. These two are computed using shared features but separate decoders shown in
Figure~\ref{fig:framework}.

\paragraph{Affine-Flow Warping.}
Let $\varphi_l^{\mathrm{o}}$ and $\varphi_l^{\mathrm{s}}$ denote the optical and SAR feature maps at level $l$. To facilitate hierarchical refinement, the SAR features are first warped using the estimated transformation from the previous scale $\hat{W}_{l}^{\mathrm{\text{prev}}}$, yielding $\tilde{\varphi}_l^{\mathrm{s}}$. The aligned pair $[\varphi_l^{\mathrm{o}}, \tilde{\varphi}_l^{\mathrm{s}}]$ is then passed to two decoders:
\begin{equation}
\hat{W}^{\mathrm{a}}_l = \mathcal{A}([\varphi_l^{\mathrm{o}}, \tilde{\varphi}_l^{\mathrm{s}}]),
\quad
\hat{W}^{\mathrm{f}}_l = \mathcal{F}([\varphi_l^{\mathrm{o}}, \tilde{\varphi}_l^{\mathrm{s}}]),
\end{equation}
where $\mathcal{A}(\cdot)$ and $\mathcal{F}(\cdot)$ denote the affine and flow regressors. The outputs $\hat{W}^{\mathrm{a}}_l,\hat{W}^{\mathrm{f}}_l$ are used to update the current transformation $\hat{W}_l$.

\paragraph{Multi-stage matching pipeline.}
The matching proceeds through three stages:

\textbf{Coarse Initialization ($l = 16$):}  
At the coarsest level, only the affine regressor is used, allowing the model to estimate an initial warp $\hat{W}_{16}$ based on global affine. This serves as a robust starting point for finer-scale refinement.

\textbf{Coupled Registration ($l = 8, 4$):}  
At intermediate levels, both affine and flow regressors are active. SAR features are warped using the previous warp \(\hat{W}_{l}^{prev}\) then concatenated with optical features. GLAM simultaneously predicts $\hat{W}^{\mathrm{a}}_l,\hat{W}^{\mathrm{f}}_l$ at this stage, then updates $\hat{W}_l$ using $\hat{W}^{\mathrm{f}}_l$. These two deformation fields are jointly optimized through a consistency loss $\mathcal{L}_{\text{cons}}$, which encourages mutual guidance between global alignment and fine-grained warping. Details of the loss are provided in the Loss Functions section.

\textbf{Refinement ($l = 2, 1$):}  
At this stage, large misalignment has been resolved, allowing the following stages to focus on fine-grained refinement. Thus, the affine regressor is omitted. Only the flow is estimated to capture residual displacements. In addition, a pixel-wise certainty map $\hat p$ is predicted as a certainty logit used for the final refinement step.

Thus, by incorporating both affine and flow transformations within a coarse-to-fine framework, GLAM achieves global structural consistency and local precision. 

\subsection{Loss Functions}
\label{sec:loss}

As part of SOMA’s hierarchical design, our loss function specifically addresses the challenges inherent in coarse-to-fine registration. In addition to the conventional RMSE-based warp loss, our optimization objective incorporates four contrapuntally designed terms to enhance certainty awareness, enforce multi-scale progressive refinement, facilitate complementary affine-flow interactions, and mitigate global drift introduced by affine transformations.

The final loss $\mathcal{L}$ is defined as:
\begin{equation}
\begin{split}
\mathcal{L} = \; & 
\underbrace{\mathcal{L}_{\mathrm{warp}}}_{\text{warp loss}}+\underbrace{\lambda\mathcal{L}_{\mathrm{cons}}}_{\text{affine-flow consistency}}\\ 
&+ \underbrace{\alpha_{\mathrm{c}} \mathcal{L}_{\mathrm{cert}}}_{\text{certainty supervision}} + \underbrace{\alpha_{\mathrm{d}} \mathcal{L}_{\mathrm{delta}}}_{\text{residual supervision}}  +
\underbrace{\alpha_{\mathrm{u}} \mathcal{L}_{\mathrm{uni}}}_{\text{patch-wise uniformity}},
\end{split}
\end{equation}

where \(\alpha_{\mathrm{c}}=\alpha_{\mathrm{d}}=\alpha_{\mathrm{u}}=0.1\), and \(\lambda=0.5\).

\paragraph{Warp Loss \(\mathcal{L}_{\mathrm{warp}}\).}
We follow standard practice by minimizing the pixel-wise RMSE between the predicted deformation field \(\hat{W}_1\) and the ground truth warp \(W\):
\begin{equation}
\mathcal{L}_{\mathrm{warp}} = \mathrm{RMSE}\left(\hat{W}_1, W_{gt}\right).
\end{equation}
This term anchors the overall registration quality. However, its performance is insufficient when faced with large modality discrepancies.

\paragraph{Affine-Flow Consistency \(\mathcal{L}_{\mathrm{cons}}\).}
GLAM simultaneously estimates affine and flow fields, with the goal of leveraging their complementary strengths. To achieve this interaction and mutual enhancement, we minimize the discrepancy between their predictions through a consistency loss function: 
\begin{equation}
\mathcal{L}_{\mathrm{cons}} =
\sum_{l \in \{8, 4\}} \mathrm{RMSE}\left(\hat{W}_l^{\mathrm{f}}, \hat{W}_l^{\mathrm{a}}\right).
\end{equation}

\paragraph{Certainty Supervision \(\mathcal{L}_{\mathrm{cert}}\).}
To address the increased modality differences arising at finer scales—particularly evident as the resolution approaches the original input—we introduce a supervision term on the predicted certainty logits \(\hat{p}\), encouraging them to reflect the actual matching quality:

\begin{equation}
\mathcal{L}_{\mathrm{cert}} =
\frac{1}{|\Omega|} \sum_{\mathbf{x} \in \Omega}
\left( \sigma(\hat{p}(\mathbf{x})) - e^{-\|\hat{W}_{1}(\mathbf{x}) - W_{gt}(\mathbf{x})\|_2} \right)^2,
\end{equation}
where \(\Omega\) denotes the image domain and \(\sigma(\cdot)\) is the sigmoid function. 

\paragraph{Multi-Scale Residual Supervision \(\mathcal{L}_{\mathrm{delta}}\).}
To stabilize hierarchical refinement under modality-induced distortions, we supervise the residual deformation at each pyramid level \(l \in \{8, 4, 2, 1\}\), defined as the difference between ground-truth and the accumulated prediction from coarser levels:
\begin{equation}
\mathcal{L}_{\mathrm{delta}} =
\sum_{l\in \{8, 4,2,1\}} w_l \left\| \hat{W}_l - (W_{gt} - \hat{W}_{l}^{\mathrm{prev}}) \right\|_1,
\end{equation}
with weights \(w_8=0.125, w_4=0.25, w_2=0.5, w_1=1\). 
$\hat{W}_l$ and $\hat{W}_{l}^{\mathrm{prev}}$ are upsampled to the size of $W_{gt}$.
\paragraph{Patch-Wise Uniformity \(\mathcal{L}_{\mathrm{uni}}\).}
Although affine transformations offer valuable global guidance for modeling coarse alignment, in some cases involving complex local distortions, they can inadvertently introduce spatial bias across certain regions. To mitigate this effect and ensure more balanced local accuracy, we divide the predicted warp \(\hat{W}_{1}\) and the ground truth \(W_{gt}\) into four non-overlapping patches $\{\hat{W}_1^i,W_{gt}^i\},i \in \{1,2,3,4\}$, and minimize the standard deviation (Std) of patch-wise registration error:
\begin{equation}
\mathcal{L}_{\mathrm{uni}} =
\operatorname{Std}(\left\{ \mathrm{RMSE}\left(\hat{W}_1^i, W_{gt}^i\right) \right\}_{i=1}^{4}).
\end{equation}

Our loss formulation extends beyond conventional objectives by explicitly addressing the structural, hierarchical, and spatial challenges in SAR-Optical matching. 
This interplay of supervision leads to more stable and accurate convergence, as validated by our experiments.

\section{Experiments}

\subsection{Datasets}
We evaluate SOMA's registration performance on two public datasets: SEN1-2\cite{sen12} dataset and GFGE\_SO\cite{rmso} dataset.
The SEN1-2 dataset encompasses diverse land cover types at 10-meter spatial resolution. Since the original image pairs are not strictly aligned, we use the finely registered version provided by SOPatch\cite{sopatch}.
GFGE\_SO dataset contains multi-temporal, multi-satellite SAR-Optical pairs with a higher spatial resolution of 5m. Compared to SEN1-2, it presents more severe spectral discrepancies and radiometric noise, posing greater challenges for cross-modal registration.
We further test SOMA on two external datasets for generalizability analysis, which are not used during training: WHU-SEN-City\cite{whu} and OSdataset\cite{osdataset}.

To replicate realistic registration scenarios, random geometric transformations are applied to the SAR images. Specifically, following the previous protocols, we introduce translations of up to 32 pixels, scale variations of 0.2, and rotations within ±5°. In ablation, we further extend the range by increasing the maximum translation to 50 pixels and the rotation span to ±20°.

\subsection{Implementation Details}
We adopt an end-to-end training strategy, where the DINOv2 backbone is kept frozen. All other components are trained from scratch.
Training is conducted for 100 epochs with a batch size of 4, using the AdamW optimizer with a learning rate of $5\times10^{-5}$. A warm-up phase of 5 epochs is applied at the beginning of training.
All experiments and evaluations are performed on two NVIDIA RTX 4090 GPUs.

\begin{figure}[t]
    \centering
     \includegraphics[width=0.35\textwidth]{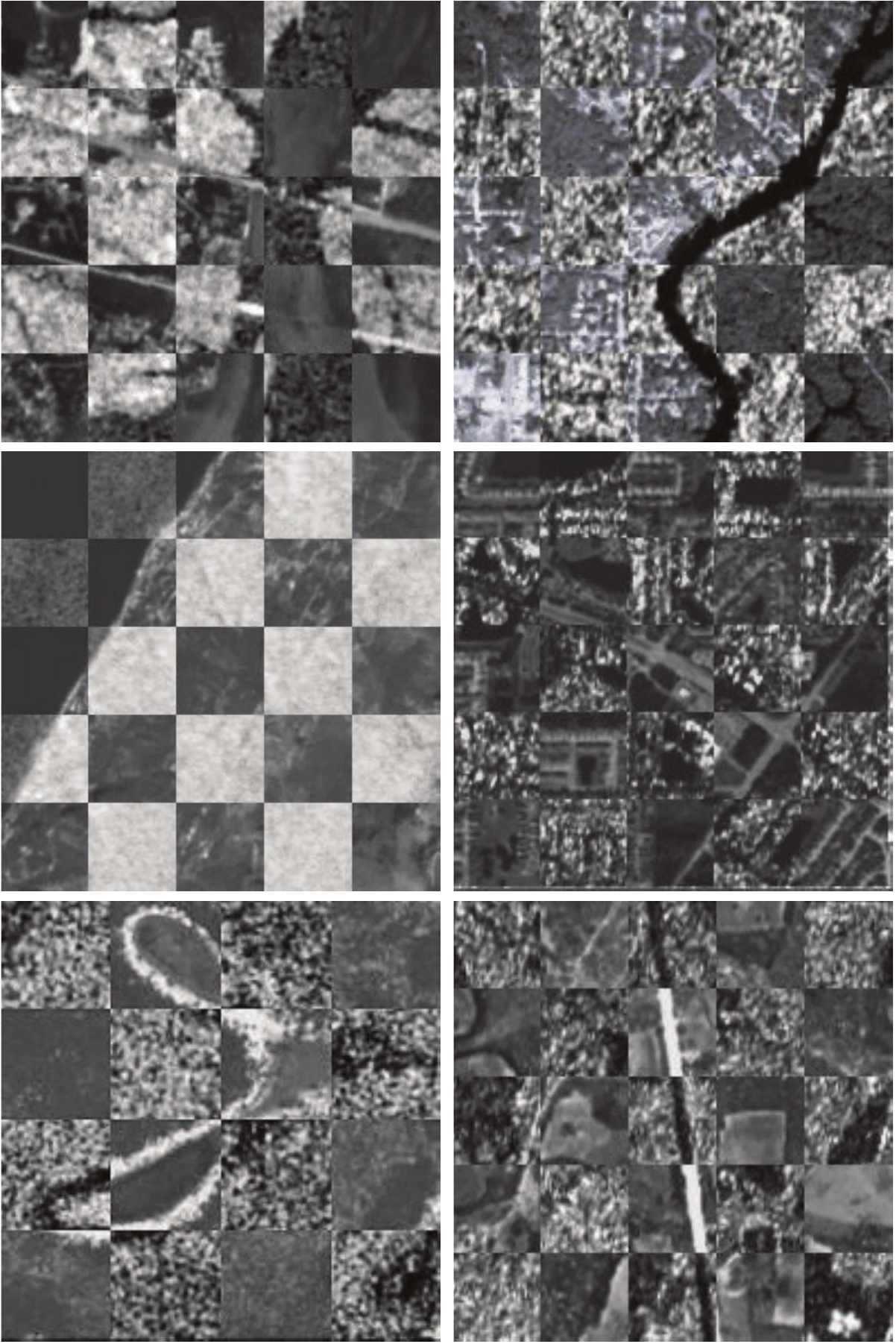}
	\caption{\textbf{Registration Results of SOMA} on the SEN1-2 (Left) and the GFGE\_SO dataset (Right).}
	\label{fig:result}
\end{figure}

\begin{table}[t]
\centering
\scriptsize
\setlength{\tabcolsep}{4pt}
\begin{tabular}{lccccc}
\toprule
\textbf{Method} & \multicolumn{5}{c}{\textbf{CMR@Threshold (\%)} $\uparrow$}  \\
\cmidrule(lr){2-6}
& 1px & 2px & 3px & 4px & 5px \\
\midrule
MI            & 44.03 & 53.45 & 57.16 & 58.78 & 60.01 \\
\cite{mi}\\
CFOG       & 46.02 & 59.73 & 64.87 & 66.58 & 67.53 \\
\cite{cfog}\\ 
DDFN          & 49.64 & 65.63 & 72.76 & 74.95 & 76.00 \\
\cite{zhang2020ddfn}\\
FFT U-Net     & 54.87 & 73.43 & 79.90 & 83.52 & 86.56 \\
\cite{fft}\\
MoPSI         & 60.87 & 81.71 & 90.18 & 92.27 & 93.60 \\
\cite{liu2023optical}\\
OSMNet       & 62.58 & 83.04 & 91.13 & 93.13 & 94.46 \\
\cite{zhang2021explore}\\
RMSO-ConvNeXt & 68.67 & 86.18 & 92.46 & 94.46 & 95.60 \\
\cite{rmso}\\
SO-ConvNeXt   & \underline{74.38} & \underline{90.27} & \underline{94.36} & \underline{95.70 }& \underline{96.27} \\
\cite{rmso}\\
\midrule
\textbf{SOMA}      & \textbf{86.67} & \textbf{94.50} & \textbf{95.97} & \textbf{97.56} & \textbf{98.78} \\
\rowcolor{gray!30}\textbf{    }      & \textit{(+12.29)} & \textit{(+4.23)} & \textit{(+1.61)} & \textit{(+1.86)} & \textit{(+2.51)} \\
\bottomrule
\end{tabular}
\caption{\textbf{Comparison of correctly match rate (CMR)} at varying pixel thresholds on the SEN1-2 dataset.}
\label{tab:sen12}
\end{table}

\subsection{Comparison with State-of-the-Art Methods}
We compare SOMA against several representative methods, including traditional methods and deep learning-based methods known for their fine-grained performance. 
The traditional side includes MI~\cite{mi} and CFOG~\cite{cfog}. And, on the learning-based side, we select a set of works—DDFN~\cite{zhang2020ddfn}, FFT U-Net~\cite{fft}, OSMNet~\cite{zhang2021explore}, MoPSI~\cite{liu2023optical}, SO-ConvNeXt and RMSO-ConvNeXt~\cite{rmso}—as current performance upper bounds.

We adopt the Correctly Matching Rate (CMR) as a primary evaluation metric to assess the robustness and precision of registration methods. CMR computes the proportion of image pairs whose registration error remains below a predefined threshold, thereby reflecting the method’s ability to consistently achieve acceptable alignment. 

Table~\ref{tab:sen12} summarizes the CMR at pixel thresholds ranging from 1-5px on the SEN1-2 dataset. Our proposed SOMA consistently achieves the best performance across all thresholds. At the most stringent criterion, SOMA attains 86.67\%, surpassing the previous state-of-the-art method by 12.29\%. This advantage persists as the matching threshold relaxes, with SOMA reaching 98.78\% at 5px, outperforming by 2.51\%. As shown in Table~\ref{tab:gfgedataset}, on the more challenging GFGE\_SO dataset, which features higher-resolution images and stronger spectral discrepancies, SOMA also surpasses the strongest baseline across all thresholds, with improvements peaking at 18.50\% under the 1px threshold.

The consistent and significant improvements across all thresholds demonstrate SOMA’s superior performance in pixel-level registration. Some examples of the results are shown in Figure~\ref{fig:result}.

\begin{table}[t]
\centering
\scriptsize
\setlength{\tabcolsep}{4pt}
\begin{tabular}{lccccc}
\toprule
\textbf{Method} & \multicolumn{5}{c}{\textbf{CMR@Threshold (\%) $\uparrow$}}  \\
\cmidrule(lr){2-6}
& 1px & 2px & 3px & 4px & 5px \\
\midrule
MI & 30.21 & 40.36 & 44.25 & 46.74 & 48.33 \\
\cite{mi}\\
CFOG & 37.98 & 46.24 & 55.01 & 59.18 & 61.87 \\
\cite{cfog}\\
DDFN & 43.05 & 52.31 & 59.68 & 64.26 & 67.05 \\
\cite{zhang2020ddfn}\\
FFT U-Net & 47.93 & 57.49 & 63.07 & 67.05 & 71.23 \\
\cite{fft}\\
MoPSI & 54.97 & 69.94 & 75.02 & 82.88 & 89.16 \\
\cite{liu2023optical}\\
OSMNet & 56.40 & 72.53 & 79.00 & 87.26 & 91.84 \\
\cite{zhang2021explore}\\
SO-ConvNeXt & 58.69 & 75.51 & 82.78 & 92.05 & 95.83 \\
\cite{rmso}\\
RMSO-ConvNeXt & \underline{60.88} & \underline{79.00} & \underline{85.37} & \underline{92.94} & \underline{96.23} \\
\cite{rmso}\\
\midrule
\textbf{SOMA} & \textbf{79.38} & \textbf{88.82} & \textbf{93.97} & \textbf{96.68} & \textbf{98.42} \\
\rowcolor{gray!30}\textbf{    }      & \textit{(+18.50)} & \textit{(+9.82)} & \textit{(+8.60)} & \textit{(+3.74)} & \textit{(+2.19)} \\
\bottomrule
\end{tabular}
\caption{\textbf{Comparison of correctly match rate (CMR)} at varying pixel thresholds on the GFGE\_SO dataset.}
\label{tab:gfgedataset}
\end{table}

\subsection{Ablation Study}

We perform an ablation study on the SEN1-2 dataset by progressively incorporating the DINOv2, FGE, and GLAM modules into a shared baseline. As shown in Table~\ref{tab:ablation}, we report results in terms of CMR, as well as the registration error $R_{\text{avg}}$, calculated as the mean RMSE across all test pairs.

Individually, the FGE module produces the most substantial improvement over the baseline, particularly under strict alignment criteria. At the 1px threshold, it increases CMR by 20.54\% compared to the baseline, demonstrating the effectiveness of feature gradient enhancement in improving fine-level characteristics. The GLAM module also contributes consistent gains, especially at 1-3px, indicating the value of coarse-to-fine affine-flow modeling in capturing hierarchical geometric transformations.

When used in combination, the benefits of each module become more pronounced. FGE + GLAM already achieves strong performance, outperforming all single-module variants. Adding DINOv2 leads to further gains, confirming the advantage of semantic regularization via frozen fundamental features. The complete framework, SOMA, integrates all three components and achieves the best results, with a CMR of 87.53\% at 1px and the lowest registration error. Further analyzes are provided in the Supplement.


\begin{table}[t]
\centering
\scriptsize
\setlength{\tabcolsep}{3pt}
\begin{tabular}{lccccc c}
\toprule
\textbf{Setup} & \multicolumn{5}{c}{\textbf{CMR@Threshold (\%)$\uparrow$}} & \textbf{R$_{avg} \downarrow $} \\
\cmidrule(lr){2-6}
& 1px & 2px & 3px & 4px & 5px & \\
\midrule
\textbf{baseline }        & \underline{28.85} & \underline{76.28} & \underline{86.19} & \underline{90.83} & \underline{93.40} & \underline{2.58} \\
\midrule
+ DINO           & 27.75 & 81.05 & 89.24 & 94.38 & 96.21 & 2.36 \\
\rowcolor{gray!15}\textbf{}        & \textit{(-1.10)} & \textit{(+4.77)} & \textit{(+3.05)} & \textit{(+3.55)} & \textit{(+2.81)} & \textit{(-0.22)} \\
+ FGE            & 49.39 & 86.31 & 91.69 & 94.50 & 97.31 & 1.95 \\
\rowcolor{gray!15}\textbf{}        & \textit{(+20.54)} & \textit{(+10.03)} & \textit{(+5.50)} & \textit{(+3.67)} & \textit{(+3.91)} & \textit{(-0.63)} \\
+ GLAM           & 42.79 & 71.15 & 84.23 & 88.75 & 92.91 & 2.25 \\
\rowcolor{gray!15}\rowcolor{gray!20}\rowcolor{gray!20}\rowcolor{gray!20}\textbf{}        & \textit{(+13.94)} & \textit{(-5.13)} & \textit{(-1.96)} & \textit{(-2.08)} & \textit{(-0.49)} & \textit{(-0.33)} \\
+ DINO + FGE     & 52.44 & 88.88 & 94.01 & 95.23 & 96.70 & 1.77 \\
\rowcolor{gray!15}\textbf{} & \textit{(+23.59)} & \textit{(+12.60)} & \textit{(+7.82)} & \textit{(+4.40)} & \textit{(+3.30)} & \textit{(-0.81)} \\
+ DINO + GLAM    & 60.88 & 84.23 & 91.81 & 95.84 & 97.92 & 1.55 \\
\rowcolor{gray!15}\rowcolor{gray!20}\textbf{}        & \textit{(+32.03)} & \textit{(+7.95)} & \textit{(+5.62)} & \textit{(+5.01)} & \textit{(+4.52)} & \textit{(-1.03)} \\
+ FGE + GLAM     & 75.43 & 87.16 & 92.91 & 95.48 & 96.70 & 1.33 \\
\rowcolor{gray!15}\textbf{}        & \textit{(+46.58)} & \textit{(+10.88)} & \textit{(+6.72)} & \textit{(+4.65)} & \textit{(+3.30)} & \textit{(-1.25)} \\
\midrule
\textbf{SOMA (full)} & \textbf{87.53} & \textbf{94.87} & \textbf{96.45} & \textbf{97.19} & \textbf{97.80} & \textbf{0.94} \\
\rowcolor{gray!30}\textbf{} & \textit{(+58.68)} & \textit{(+18.59)} & \textit{(+10.26)} & \textit{(+6.36)} & \textit{(+4.40)} & \textit{(-1.64)} \\

\bottomrule
\end{tabular}
\caption{\textbf{Component analysis of SOMA}.}
\label{tab:ablation}
\end{table}

\subsection{Generalizability Analysis}

To evaluate the generalizability of SOMA, we perform cross-dataset testing using models trained solely on SEN1-2. Two additional datasets are used for evaluation. The first, WHU-SEN-City\cite{whu}, is collected from the same satellite and shares the same spatial resolution as SEN1-2, but covers 32 cities in China, exhibiting different geographical distributions. The second, OSdataset\cite{osdataset}, is constructed from Google Earth optical images and GaoFen-3 SAR data, which feature significantly higher spatial resolution and diverse image characteristics of multiple sources. Here, we use image pairs provided by SOPatch for both datasets, consistent with the SEN1-2 setting.

The experimental results in Table~\ref{tab:generalization} show that SOMA maintains strong performance on WHU-SEN-City, indicating good generalization across scenes with similar resolution and sensor characteristics, but varying geographic content. On OSdataset, despite the lack of high-resolution training supervision, SOMA still shows reasonably good performance at 5px. This suggests that SOMA retains generalizability under cross-sensor and high-resolution domain shifts, even without explicit adaptation.

\begin{table}[t]
\centering
\scriptsize
\setlength{\tabcolsep}{4pt}
\begin{tabular}{lcccc}
\toprule
\textbf{Dataset} & \textbf{Resolution} & \textbf{Source} & \textbf{CMR@5px (\%)} & \textbf{R$_{avg}$} \\
\midrule
\textbf{WHU-SEN-City} & 10m & \makecell{Sentinel-1 \\ Sentinel-2} & 96.75 & 1.80 \\
\textbf{OSdataset} & 1m & \makecell{Google Earth \\ GaoFen-3} & 92.96 & 3.37 \\
\bottomrule
\end{tabular}
\caption{\textbf{Generalization performance of SOMA}.}
\label{tab:generalization}
\end{table}

\subsection{Runtime Analysis}
We evaluated the runtime of SOMA in comparison to the baseline. SOMA processes each image pair with a size of $512\times512$ pixels in 94 ms on average, incurring just a 6.8\% overhead compared to the baseline’s 88 ms, while delivering significantly improved accuracy.

\section{Conclusion}

We have proposed SOMA, a high-precision and robust framework for SAR-optical image registration. SOMA employs the Feature Gradient Enhancer (FGE) to refine deep features, guiding the model to focus on structural cues that are more conducive to establishing accurate matching. A frozen DINOv2 further stabilize coarse alignment. The Global-Local Affine-Flow Matcher (GLAM) jointly predicts affine and flow fields, enabling mutual guidance between global and local transformations and achieving accurate dense matching in a coarse-to-fine manner. Experimental results demonstrate that SOMA delivers significant improvements in pixel-level accuracy on different datasets and generalizes well across varying scenarios.

\section{Acknowledgments}
This work was supported by the National Natural Science Foundation of China under Grant Nos. 62476220 and 61971356, as well as the Natural Science Basic Research Program of Shaanxi Province under Grant No. 2024JC-DXWT-07.

\bibliography{aaai2026}

@article{jiang2021review,
  title={A review of multimodal image matching: Methods and applications},
  author={Jiang, Xingyu and Ma, Jiayi and Xiao, Guobao and Shao, Zhenfeng and Guo, Xiaojie},
  journal={Information Fusion},
  volume={73},
  pages={22--71},
  year={2021},
  publisher={Elsevier}
}

@article{xiang2018sift,
  title={OS-SIFT: A robust SIFT-like algorithm for high-resolution optical-to-SAR image registration in suburban areas},
  author={Xiang, Yuming and Wang, Feng and You, Hongjian},
  journal={IEEE Transactions on Geoscience and Remote Sensing},
  volume={56},
  number={6},
  pages={3078--3090},
  year={2018},
  publisher={IEEE}
}

@article{zhang2020ddfn,
  title={Optical and SAR image matching using pixelwise deep dense features},
  author={Zhang, Han and Lei, Lin and Ni, Weiping and Tang, Tao and Wu, Junzheng and Xiang, Deliang and Kuang, Gangyao},
  journal={IEEE Geoscience and Remote Sensing Letters},
  volume={19},
  pages={1--5},
  year={2020},
  publisher={IEEE}
}

@inproceedings{roma,
  title={RoMa: Robust dense feature matching},
  author={Edstedt, Johan and Sun, Qiyu and B{\"o}kman, Georg and Wadenb{\"a}ck, M{\aa}rten and Felsberg, Michael},
  booktitle={Proceedings of the IEEE/CVF Conference on Computer Vision and Pattern Recognition},
  pages={19790--19800},
  year={2024}
}

@article{dinov2,
  TITLE = {{DINOv2: Learning Robust Visual Features without Supervision}},
  AUTHOR = {Oquab, Maxime and Darcet, Timoth{\'e}e and Moutakanni, Th{\'e}o and Vo, Huy and Szafraniec, Marc and Khalidov, Vasil and Fernandez, Pierre and Haziza, Daniel and Massa, Francisco and El-Nouby, Alaaeldin and Assran, Mahmoud and Ballas, Nicolas and Galuba, Wojciech and Howes, Russell and Huang, Po-Yao and Li, Shang-Wen and Misra, Ishan and Rabbat, Michael and Sharma, Vasu and Synnaeve, Gabriel and Xu, Hu and Jegou, Herv{\'e} and Mairal, Julien and Labatut, Patrick and Joulin, Armand and Bojanowski, Piotr},
  URL = {https://hal.science/hal-04376640},
  JOURNAL = {{Transactions on Machine Learning Research Journal}},
  PUBLISHER = {{[Amherst Massachusetts]: OpenReview.net, 2022}},
  PAGES = {1-31},
  YEAR = {2024}
}

@inproceedings{dinov2reg,
  TITLE = {{Vision Transformers Need Registers}},
  AUTHOR = {Darcet, Timoth{\'e}e and Oquab, Maxime and Mairal, Julien and Bojanowski, Piotr},
  URL = {https://inria.hal.science/hal-04394066},
  BOOKTITLE = {{International Conference on Learning Representations (ICLR)}},
  YEAR = {2024},
  HAL_ID = {hal-04394066},
  HAL_VERSION = {v1},
}

@inproceedings{dinoisbest,
  title={Do computer vision foundation models learn the low-level characteristics of the human visual system?},
  author={Cai, Yancheng and Yin, Fei and Hammou, Dounia and Mantiuk, Rafal},
  booktitle={Proceedings of the Computer Vision and Pattern Recognition Conference},
  pages={20039--20048},
  year={2025}
}

@article{li2023multimodal,
  title={Multimodal image fusion framework for end-to-end remote sensing image registration},
  author={Li, Liangzhi and Han, Ling and Ding, Mingtao and Cao, Hongye},
  journal={IEEE Transactions on Geoscience and Remote Sensing},
  volume={61},
  pages={1--14},
  year={2023},
  publisher={IEEE}
}

@article{sarsift,
  title={SAR-SIFT: a SIFT-like algorithm for SAR images},
  author={Dellinger, Flora and Delon, Julie and Gousseau, Yann and Michel, Julien and Tupin, Florence},
  journal={IEEE Transactions on Geoscience and Remote Sensing},
  volume={53},
  number={1},
  pages={453--466},
  year={2014},
  publisher={IEEE}
}

@article{CoFSM,
  title={Multi-modal remote sensing image matching considering co-occurrence filter},
  author={Yao, Yongxiang and Zhang, Yongjun and Wan, Yi and Liu, Xinyi and Yan, Xiaohu and Li, Jiayuan},
  journal={IEEE Transactions on Image Processing},
  volume={31},
  pages={2584--2597},
  year={2022},
  publisher={IEEE}
}

@article{hoss,
  title={Illumination-robust remote sensing image matching based on oriented self-similarity},
  author={Sedaghat, Amin and Mohammadi, Nazila},
  journal={ISPRS Journal of Photogrammetry and Remote Sensing},
  volume={153},
  pages={21--35},
  year={2019},
  publisher={Elsevier}
}

@article{mlss,
  title={Max-index based local self-similarity descriptor for robust multi-modal image registration},
  author={Hong, Yameng and Leng, Chengcai and Zhang, Xinyue and Peng, Jinye and Jiao, Licheng and Basu, Anup},
  journal={IEEE Geoscience and Remote Sensing Letters},
  volume={19},
  pages={1--5},
  year={2022},
  publisher={IEEE}
}

@article{liu2023optical,
  title={Optical and SAR images matching based on phase structure convolutional features},
  author={Liu, Yang and Qi, Hua and Peng, Shiyong},
  journal={IEEE Geoscience and Remote Sensing Letters},
  volume={20},
  pages={1--5},
  year={2023},
  publisher={IEEE}
}

@article{zhang2021explore,
  title={Explore better network framework for high-resolution optical and SAR image matching},
  author={Zhang, Han and Lei, Lin and Ni, Weiping and Tang, Tao and Wu, Junzheng and Xiang, Deliang and Kuang, Gangyao},
  journal={IEEE Transactions on Geoscience and Remote Sensing},
  volume={60},
  pages={1--18},
  year={2021},
  publisher={IEEE}
}

@article{munet,
  title={A multiscale framework with unsupervised learning for remote sensing image registration},
  author={Ye, Yuanxin and Tang, Tengfeng and Zhu, Bai and Yang, Chao and Li, Bo and Hao, Siyuan},
  journal={IEEE Transactions on Geoscience and Remote Sensing},
  volume={60},
  pages={1--15},
  year={2022},
  publisher={IEEE}
}

@ARTICLE{zhang2024robust,
  author={Zhang, Xiaoting and Wang, Yinghua and Liu, Jun and Wang, Siyuan and Zhang, Chen and Liu, Hongwei},
  journal={IEEE Transactions on Geoscience and Remote Sensing}, 
  title={Robust Coarse-to-Fine Registration Algorithm for Optical and SAR Images Based on Two Novel Multiscale and Multidirectional Features}, 
  year={2024},
  volume={62},
  number={},
  pages={1-26},
  doi={10.1109/TGRS.2024.3417217}}

@inproceedings{clip,
  title={Learning transferable visual models from natural language supervision},
  author={Radford, Alec and Kim, Jong Wook and Hallacy, Chris and Ramesh, Aditya and Goh, Gabriel and Agarwal, Sandhini and Sastry, Girish and Askell, Amanda and Mishkin, Pamela and Clark, Jack and others},
  booktitle={International conference on machine learning},
  pages={8748--8763},
  year={2021},
}

@article{zhou2021ibot,
  title={iBOT: Image BERT Pre-Training with Online Tokenizer},
  author={Zhou, Jinghao and Wei, Chen and Wang, Huiyu and Shen, Wei and Xie, Cihang and Yuille, Alan and Kong, Tao},
  journal={International Conference on Learning Representations (ICLR)},
  year={2022}
}

@InProceedings{omniglue,
    author    = {Jiang, Hanwen and Karpur, Arjun and Cao, Bingyi and Huang, Qixing and Araujo, Andr\'e},
    title     = {OmniGlue: Generalizable Feature Matching with Foundation Model Guidance},
    booktitle = {Proceedings of the IEEE/CVF Conference on Computer Vision and Pattern Recognition (CVPR)},
    month     = {June},
    year      = {2024},
    pages     = {19865-19875}
}

@article{dong2024changeclip,
  title={ChangeCLIP: Remote sensing change detection with multimodal vision-language representation learning},
  author={Dong, Sijun and Wang, Libo and Du, Bo and Meng, Xiaoliang},
  journal={ISPRS Journal of Photogrammetry and Remote Sensing},
  volume={208},
  pages={53--69},
  year={2024},
  publisher={Elsevier}
}

@inproceedings{melekhov2019dgc,
  title={Dgc-net: Dense geometric correspondence network},
  author={Melekhov, Iaroslav and Tiulpin, Aleksei and Sattler, Torsten and Pollefeys, Marc and Rahtu, Esa and Kannala, Juho},
  booktitle={2019 IEEE Winter Conference on Applications of Computer Vision (WACV)},
  pages={1034--1042},
  year={2019}
}

@inproceedings{glunet,
  title={GLU-Net: Global-local universal network for dense flow and correspondences},
  author={Truong, Prune and Danelljan, Martin and Timofte, Radu},
  booktitle={Proceedings of the IEEE/CVF conference on computer vision and pattern recognition},
  pages={6258--6268},
  year={2020}
}

@article{murf,
  title={Murf: Mutually reinforcing multi-modal image registration and fusion},
  author={Xu, Han and Yuan, Jiteng and Ma, Jiayi},
  journal={IEEE transactions on pattern analysis and machine intelligence},
  volume={45},
  number={10},
  pages={12148--12166},
  year={2023},
  publisher={IEEE}
}

@InProceedings{SCC,
    author    = {Li, Jiafeng and Wen, Ying and He, Lianghua},
    title     = {SCConv: Spatial and Channel Reconstruction Convolution for Feature Redundancy},
    booktitle = {Proceedings of the IEEE/CVF Conference on Computer Vision and Pattern Recognition (CVPR)},
    month     = {June},
    year      = {2023},
    pages     = {6153-6162}
}

@inproceedings{aliased,
  title={How convolutional neural networks deal with aliasing},
  author={Ribeiro, Ant{\^o}nio H and Sch{\"o}n, Thomas B},
  booktitle={ICASSP 2021-2021 IEEE International Conference on Acoustics, Speech and Signal Processing (ICASSP)},
  pages={2755--2759},
  year={2021}
}

@article{sen12,
  title={The SEN1-2 dataset for deep learning in SAR-optical data fusion},
  author={Schmitt, M and Hughes, LH and Zhu, XX},
  journal={ISPRS Annals of the Photogrammetry, Remote Sensing and Spatial Information Sciences},
  volume={4},
  pages={141--146},
  year={2018},
  publisher={Copernicus Publications G{\"o}ttingen, Germany}
}

@ARTICLE{rmso,
  author={Yang, Chao and Gong, Guoqing and Liu, Chang and Deng, Jiwei and Ye, Yuanxin},
  journal={IEEE Transactions on Geoscience and Remote Sensing}, 
  title={RMSO-ConvNeXt: A Lightweight CNN Network for Robust SAR and Optical Image Matching Under Strong Noise Interference}, 
  year={2025},
  volume={63},
  number={},
  pages={1-13},
  doi={10.1109/TGRS.2025.3550936}}

@article{sopatch,
  title={SAR-optical feature matching: A large-scale patch dataset and a deep local descriptor},
  author={Xu, Wangyi and Yuan, Xinhui and Hu, Qingwu and Li, Jiayuan},
  journal={International Journal of Applied Earth Observation and Geoinformation},
  volume={122},
  pages={103433},
  year={2023},
  publisher={Elsevier}
}

@article{fft,
  title={SAR-optical image matching by integrating Siamese U-Net with FFT correlation},
  author={Fang, Yuyuan and Hu, Jun and Du, Chuan and Liu, Zhibo and Zhang, Lei},
  journal={IEEE Geoscience and Remote Sensing Letters},
  volume={19},
  pages={1--5},
  year={2021},
  publisher={IEEE}
}

@article{cfog,
  title={Fast and robust matching for multimodal remote sensing image registration},
  author={Ye, Yuanxin and Bruzzone, Lorenzo and Shan, Jie and Bovolo, Francesca and Zhu, Qing},
  journal={IEEE Transactions on Geoscience and Remote Sensing},
  volume={57},
  number={11},
  pages={9059--9070},
  year={2019},
  publisher={IEEE}
}

@article{mi,
  title={Mutual-information-based registration of TerraSAR-X and Ikonos imagery in urban areas},
  author={Suri, Sahil and Reinartz, Peter},
  journal={IEEE Transactions on Geoscience and Remote Sensing},
  volume={48},
  number={2},
  pages={939--949},
  year={2009},
  publisher={IEEE}
}

@ARTICLE{whu,
  author={Wang, Lei and Xu, Xin and Yu, Yue and Yang, Rui and Gui, Rong and Xu, Zhaozhuo and Pu, Fangling},
  journal={IEEE Access}, 
  title={SAR-to-Optical Image Translation Using Supervised Cycle-Consistent Adversarial Networks}, 
  year={2019},
  volume={7},
  number={},
  pages={129136-129149},
  doi={10.1109/ACCESS.2019.2939649}}

@ARTICLE{osdataset,
  author={Xiang, Yuming and Tao, Rongshu and Wang, Feng and You, Hongjian and Han, Bing},
  journal={IEEE Journal of Selected Topics in Applied Earth Observations and Remote Sensing}, 
  title={Automatic Registration of Optical and SAR Images Via Improved Phase Congruency Model}, 
  year={2020},
  volume={13},
  number={},
  pages={5847-5861},
  doi={10.1109/JSTARS.2020.3026162}}

@Article{deepdes,
AUTHOR = {Dong, Yunyun and Jiao, Weili and Long, Tengfei and Liu, Lanfa and He, Guojin and Gong, Chengjuan and Guo, Yantao},
TITLE = {Local Deep Descriptor for Remote Sensing Image Feature Matching},
JOURNAL = {Remote Sensing},
VOLUME = {11},
YEAR = {2019},
NUMBER = {4},
ARTICLE-NUMBER = {430},
URL = {https://www.mdpi.com/2072-4292/11/4/430},
ISSN = {2072-4292},
DOI = {10.3390/rs11040430}
}

@ARTICLE{deepfea,
  author={Quan, Dou and Wang, Shuang and Gu, Yu and Lei, Ruiqi and Yang, Bowu and Wei, Shaowei and Hou, Biao and Jiao, Licheng},
  journal={IEEE Transactions on Geoscience and Remote Sensing}, 
  title={Deep Feature Correlation Learning for Multi-Modal Remote Sensing Image Registration}, 
  year={2022},
  volume={60},
  number={},
  pages={1-16},
  doi={10.1109/TGRS.2022.3187015}}

@InProceedings{Xu_2022_CVPR,
    author    = {Xu, Han and Ma, Jiayi and Yuan, Jiteng and Le, Zhuliang and Liu, Wei},
    title     = {RFNet: Unsupervised Network for Mutually Reinforcing Multi-Modal Image Registration and Fusion},
    booktitle = {Proceedings of the IEEE/CVF Conference on Computer Vision and Pattern Recognition (CVPR)},
    month     = {June},
    year      = {2022},
    pages     = {19679-19688}
}

@article{affine,
title = {Learning affine transformations},
journal = {Pattern Recognition},
volume = {32},
number = {10},
pages = {1783-1799},
year = {1999},
issn = {0031-3203},
author = {George Bebis and Michael Georgiopoulos and Niels {da Vitoria Lobo} and Mubarak Shah}
}

@ARTICLE{psic,
  author={Zhang, Xiaoting and Wang, Yinghua and Liu, Hongwei},
  journal={IEEE Geoscience and Remote Sensing Letters}, 
  title={Robust Optical and SAR Image Registration Based on OS-SIFT and Cascaded Sample Consensus}, 
  year={2022},
  volume={19},
  number={},
  pages={1-5},
  doi={10.1109/LGRS.2021.3069761}}

@inproceedings{cbam,
  title={Cbam: Convolutional block attention module},
  author={Woo, Sanghyun and Park, Jongchan and Lee, Joon-Young and Kweon, In So},
  booktitle={Proceedings of the European conference on computer vision (ECCV)},
  pages={3--19},
  year={2018}
}



\section*{Supplementary material}

# Supplementary Material of SOMA.

Here we further provide more detail about SOMA.

### A. FURTHER DETAILS OF FEATURE EXTRACTION

We extract a 5-level feature pyramid from key stages of the ResNet50 and DINOv2.

Specifically, features are extracted from scale levels $l \in \{1, 2, 4, 8\}$, with corresponding feature dimensions $\{3, 64, 256, 512\}$. Then, feature maps at $l = 16$ with channel dimension 1024 are from a frozen DINOv2 with register(ViT-L/14 distilled version).

### B. FURTHER DETIALS OF EXPERIMENTAL SETTINGS

Our main experiments are conducted on finely aligned datasets. To replicate realistic registration scenarios, random geometric transformations, including translation, scaling, and rotation, are applied to the SAR images. Specifically, following the protocols adopted by previous methods, we introduce translations of up to 32 pixels, scale variations of 0.2, and rotations within ±5°. To demonstrate the significant improvement brought by our module in challenging high-precision registration tasks, we further extend the deformation range by increasing the maximum translation to 50 pixels and the rotation span to ±20°, establishing a more challenging benchmark for ablation.

### C. FURTHER DETAILS ABOUT FGE SETTINGS

In the design of the FGE module, we ultimately adopt a set of $N = 8$ rotated Sobel-like filter branches. This configuration is the result of a trade-off between registration accuracy and computational efficiency. To further justify this setting, we evaluate different directional configurations on the SEN1-2 dataset, including:

- $N = 2$: {0°, 90°}
- $N = 4$: {0°, 45°, 90°, 135°}
- $N = 8$: {0°, 22.5°, 45°, ..., 157.5°}
- $N = 12$: {0°, 15°, 30°, ..., 165°}

As shown in Fig S1, more directional branches improve registration performance, reflecting an enhancement in structural awareness. However, the performance gain plateaus between $N = 8$ and $N = 12$, indicating a clear marginal effect. Meanwhile, under our experimental environment, each additional two branches incur approximately 1–2 ms of runtime overhead. Therefore, $N = 8$ provides a practical balance between performance gain and computational cost.

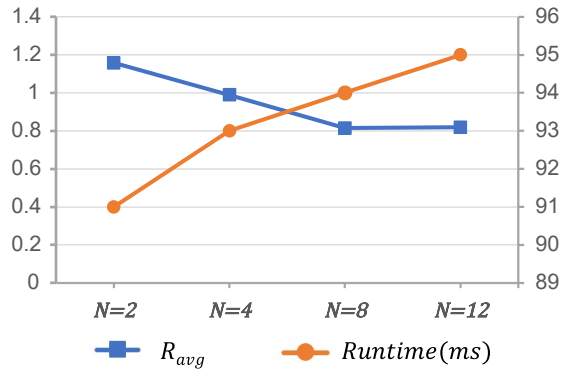

**Fig. S1.** Comparison of different gradient branch settings.

## D. FURTHER ABLATION STUDY ON GRADIENT INTEGRATION

To clarify how feature-level gradient filtering and attention cooperate in FGE, we construct five representative variants:

- **Keep Redundancy** removes SRU and CRU before gradient extration;
- **NoAttention-NoFilter** removes both directional gradient filter and attention;
- **Filter-only** keeps the fixed directional gradient filter but disables attention;
- **ConvFilter** substitutes the Sobel-like kernels with learnable $3 \times 3$ convolutions;
- **Full** retains all FGE modules.

Experimental results are summarized in Table S1. The **Keep Redundancy** variant fails to converge during training, with the loss oscillating and eventually producing NaN values. This instability highlights the critical role of reconstructing the feature space to suppress redundancy, thereby enabling effective gradient extraction.

Among the remaining four variants, the **Full** FGE achieves the best performance, consistent with our design intuition. The **NoAttention–NoFilter** variant exhibits a substantial performance drop, particularly at the finest threshold (CMR@1px = 64.18%), confirming that the synergy between directional filters and attention is an effective strategy to utilize gradient information.

To further validate the effectiveness of our gradient enhancement strategy in the feature space, we replace the fixed directional filters with learnable convolutions (**ConvFilter**). This modification leads to a noticeable drop in fine-scale accuracy. The consistently low CMR@1px observed in both the **ConvFilter** and **NoAttention–NoFilter** variants suggests that learnable convolutions alone are insufficient to capture the structural priors necessary for precise alignment.

Notably, the **Filter-only** configuration—where gradient filters are applied without attention—performs the worst among all variants, with CMR@1px plummeting to just 17.97% and the registration error nearly doubling. This indicates that simply injecting gradient information, without a mechanism to guide its integration, not only fails to help but can even degrade performance. Without attention to reweight and selectively fuse the filtered responses, the model is unable to effectively leverage the structural cues embedded in the gradients.

In summary, these results support the core design rationale of FGE: structural enhancement is not merely about adding gradient signals, but about orchestrating a coordinated process that includes redundancy reduction, directional abstraction, and attentive integration. The full FGE module effectively combines these steps to enable accurate and robust multi-modal matching.

| Setup | CMR@Threshold (%) | | | | | $R_{avg}$ |
|---|---|---|---|---|---|---|
| | 1px | 2px | 3px | 4px | 5px | |
| Keep Redundancy | - | - | - | - | - | - |
| NoAttention–NoFilter | 64.18 | 91.20 | 95.35 | 97.07 | 97.92 | 1.2700 |
| Filter-only (no attention) | 17.97 | 74.69 | 92.42 | 96.82 | 97.80 | 2.0725 |
| ConvFilter (no grad) | 67.73 | 91.32 | 95.23 | 97.43 | 98.29 | 1.2034 |
| **Full** | **86.67** | **94.50** | **95.97** | **97.56** | **98.78** | **0.9300** |

**Table S1. Ablation on the main components of FGE**. We test the effect of removing attention, gradient filters, or replacing filters with learnable convolution. The full configuration yields the best results, highlighting the importance of both gradient information and attention-based integration.

## E. MATCHABILITY ANALYSIS OF FGE

To investigate the matching capability of FGE-enhanced features, we additionally design a feature-level mapping experiment to assess how well SAR features can predict their optical counterparts. For both before and after FGE enhancement, we train two separate MLPs: one on real matched pixel-wise feature pairs and another on random mismatched pairs. Each MLP learns to predict corresponding optical features from SAR features, and the prediction is used to evaluate the capacity of features to establish reliable distinctive matching relations.

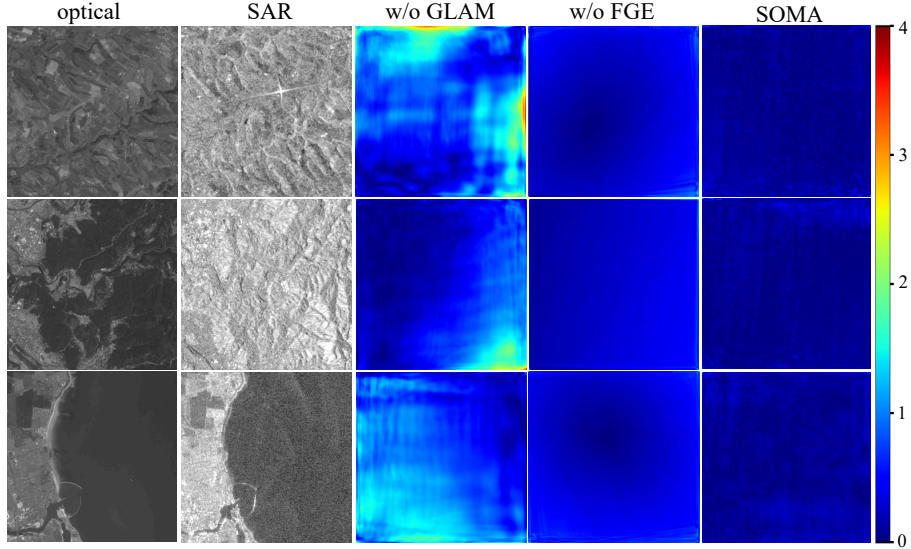

**Fig. S2.** Qualitative comparison of error heat map. Each row shows an optical–SAR image pair followed by their error maps in different settings of SOMA.

**Experimental Setup.** We randomly sample 20,000 pixel-wise feature pairs from the encoder outputs, both *before* and *after* applying the FGE module. Features are from the $l = 4$ level of the pyramids, where each feature vector has dimension $C = 128$. The architecture of the MLP used in our matchability analysis is as follow:

- **Input layer**: Linear(*C*, 128)
- **Activation**: ReLU
- **Output layer**: Linear(128, *C*)

Each MLP is trained for 20 epochs using the Adam optimizer (learning rate 1e-3, batch size 512) to minimize mean squared error (MSE) between the predicted and ground-truth optical features. Real and random pairs are trained independently, and performance is evaluated on held-out validation samples.

**Evaluation Metric.** We report the Fold-Change between random and real MSE as feature's matchability, along with 95% bootstrap confidence intervals and Mann-Whitney U test $p$-values to assess significance. Larger Fold-Change values are preferable, as they indicate a more pronounced distinction between real and random pairs

$$\text{Fold-Change} = \frac{\text{MSE}_{\text{random}}}{\text{MSE}_{\text{real}}}, \tag{S1}$$

where:

- $\text{MSE}_{\text{real}}$: MSE between predicted and ground-truth optical features for real matched SAR-optical pixel pairs.
- $\text{MSE}_{\text{random}}$: MSE between predicted and ground-truth optical features for random mismatched SAR-optical pixel pairs.

This evaluation strategy allows us to quantify how well the feature distinguishes true correspondences from random ones: a higher Fold-Change indicates that features are more predictive of true correspondences rather than random ones. The 95% confidence interval ensures the reliability of this metric across samples, while the Mann–Whitney U test assesses whether the observed difference is statistically significant and not due to chance.

**Results.** As shown in Table S2, the Fold-Change increases from 1.22× to 2.36× after applying FGE. This indicates that structural enhancement significantly improves cross-modal matchability. All results are statistically significant ($p < 10^{-5}$).

3

|  | Fold-Change | 95% CI | MWU $p$-value | Similarity |
| --- | --- | --- | --- | --- |
| Before FGE | 1.22× | [1.16, 1.27] | $1.4 \times 10^{-6}$ | 0.2409 |
| After FGE | 2.36× | [1.99, 2.79] | $4.1 \times 10^{-8}$ | 0.0132 |

**Table S2.** Matchability improvement before and after FGE.

Interestingly, despite the clear improvement in matchability, the feature similarity drops significantly—from 0.2409 to 0.0132—after FGE is applied. This phenomenon indicates that FGE does not rely on achieving modality consistency in the feature space to enable effective correspondence establishment. This counterintuitive finding challenges the common assumption that high feature similarity is essential, and instead suggests that consistent feature representations are not a necessary condition for establishing accurate and robust multi-modal correspondences.

### F. QUALITATIVE ANALYSIS

Fig S2 presents a qualitative visualization of the registration error when removing either the FGE or GLAM module.

When GLAM is removed, SOMA relies solely on the flow-based deformation field for regression. In this case, significant local inconsistencies emerge in the error map, indicating noticeable distortions in fine-grained alignment. On the other hand, when FGE is excluded, we observe a global increase in registration error compared to the full SOMA, with high-precision matching regions becoming almost imperceptible. These results demonstrate that both GLAM and FGE play indispensable roles in maintaining local precision and global structural consistency in SAR-Optical registration.

### G. CODE AVAILABILITY

The source code of SOMA and related weights will be released upon acceptance.



\end{document}